\documentclass{article}

\usepackage{microtype}
\usepackage{graphicx}
\usepackage{subcaption}
\usepackage{booktabs} 

\usepackage{hyperref}

\usepackage[preprint]{icml2026}

\usepackage{amsmath}
\usepackage{amssymb}
\usepackage{mathtools}
\usepackage{amsthm}
\usepackage{xspace}
\usepackage{multirow}
\usepackage{multicol}
\usepackage{makecell}

\usepackage[capitalize,noabbrev]{cleveref}

\theoremstyle{plain}

\theoremstyle{definition}

\theoremstyle{remark}

\newcommand{\app}{HARR\xspace}

\usepackage[textsize=tiny]{todonotes}

\icmltitlerunning{Reinforcement Fine-Tuning for History-Aware Dense Retriever in RAG}

\begin{document}
\graphicspath{{figures/}}

\twocolumn[
  \icmltitle{Reinforcement Fine-Tuning for History-Aware Dense Retriever in RAG}



  \icmlsetsymbol{equal}{*}

  \begin{icmlauthorlist}
    \icmlauthor{Yicheng Zhang}{zju}
    \icmlauthor{Zhen Qin}{zju-nb}
    \icmlauthor{Zhaomin Wu}{nus}
    \icmlauthor{Wenqi Zhang}{zju-nb}
    \icmlauthor{Shuiguang Deng}{zju}
  \end{icmlauthorlist}

  \icmlaffiliation{zju}{College of Computer Science and Technology, Zhejiang University, Hangzhou, China}
  \icmlaffiliation{zju-nb}{Ningbo Global Innovation Center, Zhejiang University, Ningbo, China}
  \icmlaffiliation{nus}{Department of Computer Science, National University of Singapore, Singapore}

  \icmlcorrespondingauthor{Zhen Qin}{zhenqin@zju.edu.cn}
  \icmlcorrespondingauthor{Shuiguang Deng}{dengsg@zju.edu.cn}

  \icmlkeywords{Machine Learning, ICML}

  \vskip 0.3in
]



\printAffiliationsAndNotice{}  

\begin{abstract}
    Retrieval-augmented generation (RAG) enables large language models (LLMs) to produce evidence-based responses, and its performance hinges on the matching between the retriever and LLMs. Retriever optimization has emerged as an efficient alternative to fine-tuning LLMs. However, existing solutions suffer from objective mismatch between retriever optimization and the goal of RAG pipeline. Reinforcement learning (RL) provides a promising solution to address this limitation, yet applying RL to retriever optimization introduces two fundamental challenges: 1) the deterministic retrieval is incompatible with RL formulations, and 2) state aliasing arises from query-only retrieval in multi-hop reasoning.
    To address these challenges, we replace deterministic retrieval with stochastic sampling and formulate RAG as a Markov decision process, making retriever optimizable by RL. Further, we incorporate retrieval history into the state at each retrieval step to mitigate state aliasing. Extensive experiments across diverse RAG pipelines, datasets, and retriever scales demonstrate consistent improvements of our approach in RAG performance.
\end{abstract}
\section{Introduction}
\label{sec:intro}

Retrieval-Augmented Generation (RAG) has become a widely adopted framework for grounding large language models (LLMs) in external knowledge by retrieving supporting evidence and conditioning generation on the retrieved context \cite{fan2024survey,lewis2020retrieval}. 
This retrieval-conditioned generation paradigm can alleviate hallucinations \cite{niu2024ragtruth} and has proven particularly beneficial for complex question answering \cite{trivedi2023interleaving}, including multi-hop reasoning that requires composing evidence across multiple sources. 
Nevertheless, the retriever and LLM-based generator are typically optimized in isolation, which can induce an objective mismatch between retrieval relevance and downstream answer correctness, thereby limiting end-to-end RAG performance.

Previous studies have investigated fine-tuning the LLM to better incorporate retrieved documents, i.e., adapting the LLMs to a given retriever.
As LLMs continue to scale, their optimization is increasingly becoming a resource-intensive task \cite{naveed2025comprehensive}.
In this context, a growing body of work instead focuses on improving the retriever to enhance RAG efficiency and effectiveness, i.e., adapting the retriever to the LLMs \cite{shi2024replug}.
Currently, retriever-centric optimization is commonly conducted via supervised fine-tuning (SFT) with synthetic or proxy targets, which may not faithfully reflect downstream answer correctness. 
As a result, the retriever can be optimized toward objectives that are only weakly aligned with the end task, limiting overall RAG performance.

To directly align retriever optimization with the end-to-end objective of RAG, we seek to optimize the retriever using reinforcement learning (RL) based on downstream task rewards; however, this setting gives rise to two key challenges: 
(1) Standard retrieval is typically realized via deterministic top-$k$ selection, which is incompatible with RL that requires a stochastic policy capable of exploring alternative actions through sampling.
(2) In multi-hop reasoning, conditioning retrieval solely on the current query leads to state ambiguity, as identical queries may arise from different retrieval histories with distinct information needs, thereby hindering effective reward assignment.


To address the above two challenges, this work proposes \app, an end-to-end reinforcement fine-tuning framework for \textbf{H}istory-\textbf{A}ware \textbf{R}einforced \textbf{R}etriever in RAG pipelines. 
\app addresses the above two challenges through two strategic designs.
To address challenge (1), we replace deterministic top-$k$ retrieval with probabilistic document sampling, allowing the retriever to be modeled as a stochastic policy that supports exploration, and then formulate retrieval-augmented question answering as a Markov decision process (MDP). These efforts make the optimization of retriever amenable to RL.
To tackle challenge (2), we incorporate retrieval history into the state at each hop, enabling the retriever to capture the evolving information need across retrieval steps and mitigating state ambiguity in multi-hop reasoning.
With these designs, \app aligns the optimization of retriever with end-to-end task rewards.

The contributions of this work are summarized as follows:
\begin{itemize}
    \item We replace the conventional deterministic top-$k$ retrieval with probabilistic document sampling, and formulate retrieval-augmented question answering as an MDP, thereby making dense retrievers amenable to RL. 
    \item We consider state aliasing as a key challenge in applying RL to optimize the retriever in multi-hop RAG and propose a history-aware state representation that conditions retrieval on the history, in order to reduce state ambiguity in multi-hop reasoning.
    \item We perform extensive experiments across diverse RAG pipelines, datasets, and retriever scales to validate the effectiveness of the proposed framework. The experimental results demonstrate that the reinforcement fine-tuning of retrievers consistently improves downstream RAG performance. Our code is available at \url{https://github.com/zyc140345/HARR}.
\end{itemize}
\section{Related Work}
\label{sec:related-work}

\subsection{RAG Workflow}
RAG systems can be categorized by their retrieval--generation workflows, which define how retrieval is invoked and how retrieved evidence is consumed during generation.
Existing workflows range from simple single-hop pipelines to more structured multi-hop and agentic designs, offering different trade-offs in reasoning capability, system complexity, and efficiency~\cite{fan2024survey, singh2025agentic}.

\paragraph{Single-Hop RAG.}
Single-hop RAG follows a retrieve--then--generate paradigm, where relevant documents are retrieved once based on the input query and directly provided to the LLM.
This workflow is widely adopted in early and practical RAG systems due to its simplicity and efficiency~\cite{lewis2020retrieval, guu2020retrieval}, and has been shown effective for factoid or shallow knowledge-intensive tasks~\cite{izacard2021leveraging}.
However, performing retrieval only once limits its ability to support complex, multi-evidence reasoning.
Accordingly, our work focuses on multi-hop and agentic settings to improve applicability.

\paragraph{Multi-Hop RAG.}
To address queries requiring compositional or multi-step reasoning, multi-hop RAG extends the workflow to perform retrieval iteratively or through structured reasoning paths.
Representative designs include iterative retrieval--generation loops~\cite{trivedi2023interleaving, shao2023enhancing} as well as tree-~\cite{yao2023tree} or graph-~\cite{besta2024graph} structured workflows that explore and aggregate multiple reasoning branches.
However, these workflows typically rely on independently pretrained components composed heuristically at inference time.
Our method addresses this limitation by optimizing the retriever to better support downstream reasoning.

\paragraph{Agentic and Adaptive Workflows.}
Recent work further generalizes multi-hop RAG into agentic workflows, where autonomous agents dynamically decide when and how to retrieve information.
These systems incorporate planning, reflection, or tool use to adapt retrieval strategies based on intermediate states~\cite{yao2023react, schick2023toolformer}, and include methods such as FLARE~\cite{jiang2023active} that trigger retrieval based on generation uncertainty.
However, retrieval decisions in agentic workflows are often governed by fixed heuristics or separately trained modules.
Our approach complements these workflows by learning retrieval policies directly optimized for downstream task performance.

Overall, existing RAG workflows improve reasoning capability through more elaborate retrieval control,
but largely rely on heuristics or fixed policies to orchestrate retrieval and generation.
This leaves open the question of how to systematically learn retrieval decisions within RAG pipelines.

\subsection{Optimization for RAG Components}
Most RAG systems are constructed by independently pretraining retrievers and LLMs with different objectives, and then integrating them into a single pipeline at inference time. This loose coupling often leads to suboptimal cooperation. Recent work seeks to reduce this mismatch by optimizing different components of the RAG pipeline.

\paragraph{Optimizing the LLMs.}
A first line of work improves the LLM's ability to plan, retrieve, and utilize evidence. Supervised fine-tuning (SFT) methods rely on annotated or synthesized trajectories to teach iterative retrieval and reasoning, such as ITER-RETGEN~\cite{shao2023enhancing}, Self-RAG~\cite{asai2024selfrag}, and CoRAG~\cite{wang2025chain}. Reinforcement learning (RL) further optimizes generation or search behaviors with outcome-based rewards, as explored in Search-R1~\cite{jin2025search}, DeepRetrieval~\cite{jiang2025deepretrieval}, and ZeroSearch~\cite{sun2025zerosearch}.
While effective, LLM-centric optimization is often expensive due to the large parameter scale of LLMs and the cost of repeated rollouts, and does not directly address cases where critical evidence is not recalled by the underlying retriever.
Our work instead focuses on lightweight retriever optimization to improve evidence recall while keeping the LLM fixed.

\paragraph{Joint Optimization.}
Another direction jointly optimizes retrievers and LLMs to improve end-to-end performance. Early work such as RAG~\cite{lewis2020retrieval} treats retrieved documents as latent variables and maximizes marginal likelihood, while subsequent methods improve optimization efficiency or differentiability through sampling-based training~\cite{zamani2024stochastic}, staged optimization~\cite{lin2024ra}, or preference-based objectives~\cite{li2024rag}.
Despite strong empirical results, joint optimization typically incurs high engineering and computational cost, and often requires both the retriever and LLM to be white-box models, which limits flexibility.
In contrast, our approach focuses on retriever-only optimization to reduce training cost and improve deployment flexibility.

\paragraph{Inserting Connecting Modules.}
Instead of updating large models, some approaches introduce lightweight modules between retrieval and generation to better mediate their interaction. Examples include external policies for retrieval control~\cite{kulkarni2024reinforcement}, multi-agent memory selection~\cite{wang2024m}, and context selection~\cite{ke2024bridging} or reconstruction~\cite{li2025oreo}. These methods are training-efficient and flexible across different backbones, but add extra system components and inference steps, and their effectiveness is constrained by the quality of the retrieved evidence.
Our work addresses this limitation by directly optimizing the retriever to influence which evidence is retrieved, without introducing additional inference modules or overhead.

\paragraph{Optimization of Retriever.}
Motivated by the limitations of LLM-centric optimization, joint training, and lightweight bridging modules, recent work explores optimizing the retriever to better balance resource cost, optimization headroom, and deployment flexibility.
Compared to LLMs and intermediate modules, retriever optimization is lightweight, can directly alter retrieved evidence, and naturally supports black-box LLMs.
Representative approaches include REPLUG~\cite{shi2024replug}, which distills LLM document preferences into the retriever, and Reward-RAG~\cite{nguyen2024reward} and R3~\cite{zhou2026optimizingretrievalragreinforcement}, which leverage answer-level feedback via contrastive or reinforcement-inspired objectives.
However, most existing methods rely on proxy objectives rather than directly optimizing downstream generation performance, leaving room for more direct retriever–generation objective matching.
Related retrieval methods outside the RAG setting also explore reinforcement learning for dense retrieval~\cite{liu2025taosearchemb} or memory selection~\cite{zhou2025memento}, but differ in task formulation or retrieval substrates, thus do not address lightweight, train–inference objective consistency within RAG pipelines.
Our work fills this gap by enabling lightweight retriever optimization that directly matches retrieval behavior to downstream generation objectives.
\section{Problem Formulation}
We consider a multi-hop retrieval-augmented generation (RAG) system comprising an LLM $\pi_{\text{LLM}}$, a dense retriever $\pi_{\text{ret}}$, and an external knowledge corpus $\mathcal{D} = \{d_i\}_{i=1}^{N}$.
The interaction is modeled as a sequential decision-making process starting with an initial query $q_0$.

At each step $t$, the LLM $\pi_{\text{LLM}}$ conditions on the accumulated retrieval history $\mathcal{H}_{t-1}$ ($\mathcal{H}_0 = (q_0)$) to determine the next output, which is either a termination signal or a sub-query $q_t$.
Given a sub-query $q_t$, the retriever $\pi_{\text{ret}}$ selects a ranked list of top-$k$ documents $D_t = (d_t^1, \ldots, d_t^k)$ from $\mathcal{D}$.
Subsequently, the LLM synthesizes an intermediate observation $o_t \sim \pi_{\text{LLM}}(\cdot \mid q_t, D_t)$ to summarize the evidence, updating the history to $\mathcal{H}_t = \mathcal{H}_{t-1} \circ (q_t, o_t)$.
Upon termination after $T$ steps, the LLM produces a final answer $y \sim \pi_{\text{LLM}}(\cdot \mid \mathcal{H}_T)$.

The overarching learning objective is to optimize the system parameters such that the generated answer $y$ maximizes a utility function $\mathrm{Score}(y, y^*)$ (e.g., F1 score or Exact Match) with respect to the ground-truth answer $y^*$. While standard approaches often optimize this via proxy losses, our goal is to align the retrieval policy $\pi_{\text{ret}}$ directly with this end-to-end generation metric.

\section{Approach} \label{sec:approach}
\begin{figure*}[t]
    \centering
    \includegraphics[width=\textwidth]{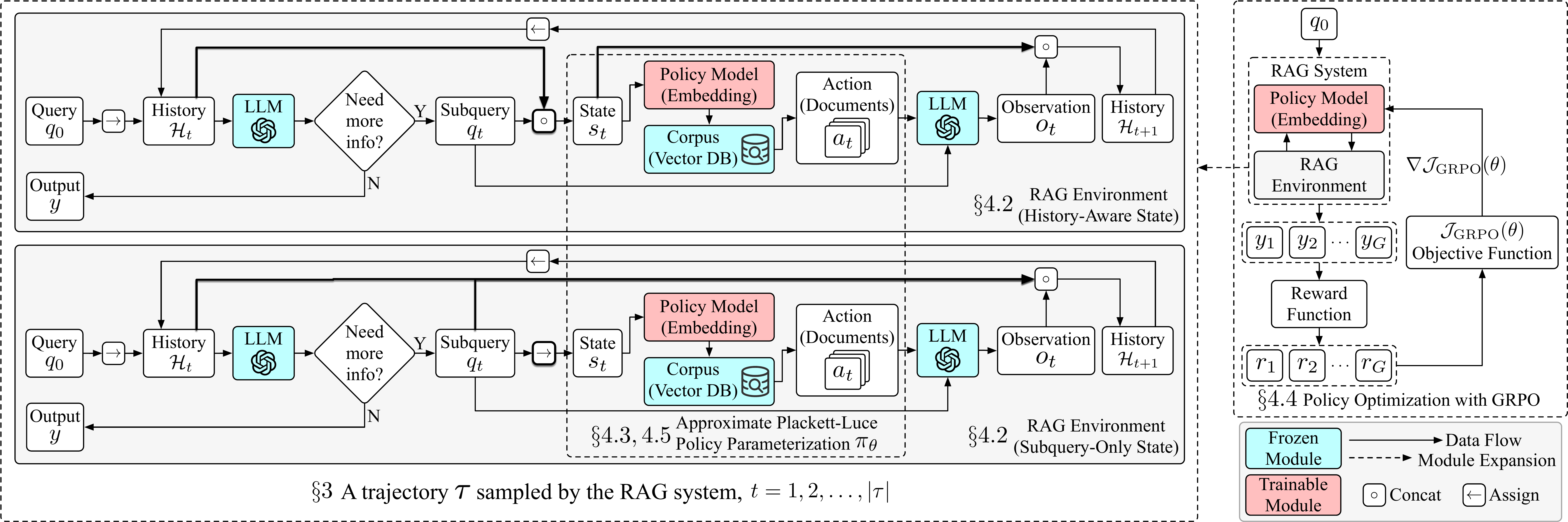}
    \caption{Overview of \app}
    \label{fig:workflow}
\end{figure*}

\subsection{Overview}
In this section, we present the proposed reinforcement learning framework for matching retriever behavior with LLMs' preferences in multi-hop RAG systems. We formulate retrieval as an MDP with a history-aware state representation to address state aliasing (\S\ref{sec:mdp_formulation}), and parameterize the retriever as a stochastic ranking policy over ordered document lists (\S\ref{sec:policy_formulation}). The retriever is optimized using Group Relative Policy Optimization (GRPO) with sparse terminal rewards (\S\ref{sec:policy_opt}), together with practical approximations that make training feasible on large-scale corpora (\S\ref{sec:approx}).

\subsection{MDP Formulation} \label{sec:mdp_formulation}
To enable RL of the retriever in multi-hop RAG systems, we formulate the interaction between the retriever and the rest of the RAG pipeline as a Markov decision process (MDP) $\mathcal{M} = (\mathcal{S}, \mathcal{A}, P, r)$. In this formulation, the retriever acts as the policy, while the LLM and the document corpus define the environment dynamics. We now specify each component of the MDP.

\paragraph{History-Aware State Space $\mathcal{S}$.}
Given the retriever as the policy, a straightforward state design defines the state solely by the current sub-query $q_t$, which matches the retriever's pretraining objective.
However, this state representation suffers from state aliasing. In multi-hop RAG, identical sub-queries $q_t$ may arise from different retrieval histories $\mathcal{H}_{t-1}$, yet be mapped to the same state under a query-only representation. Consequently, executing the same retrieval action in this apparent state can lead to different downstream reasoning trajectories and final rewards, yielding inconsistent learning signals for the same state--action pair. This leads to high-variance policy gradient estimates and hinders effective learning.

To mitigate the state aliasing issue, we define each state $s_t \in \mathcal{S}$ at step $t$ as a history-aware representation
\begin{equation}
    s_t = (\mathcal{H}_{t-1}, q_t)\text{,}
\end{equation}
where $\mathcal{H}_{t-1}$ denotes the retrieval history up to step $t-1$. 
By incorporating the retrieval history into the state, retrieval decisions made under identical sub-queries but different histories are distinguished. As a result, state-action pairs that would otherwise be aliased under a query-only state formulation are separated, leading to more consistent downstream outcomes and learning signals for each state--action pair.

\paragraph{Action Space $\mathcal{A}$.}
At each step $t$, the action is to select a ranked list of $k$ documents from the corpus $\mathcal{D}$, denoted $a_t = D_t = (d_t^1, \ldots, d_t^k)$. Thus, the action space $\mathcal{A}$ comprises all such ordered $k$-documents, with $|\mathcal{A}| = \frac{|\mathcal{D}|!}{(|\mathcal{D}| - k)!}$.

\paragraph{State Transition $P$.}
Given state $s_t = (\mathcal{H}_{t-1}, q_t)$ and action $a_t = D_t$, the state transition is induced by the LLM. Conditioned on the current sub-query and retrieved documents, the LLM produces an intermediate observation
\begin{equation}
    o_t \sim \pi_{\text{LLM}}(\cdot \mid q_t, D_t)\text{,}
\end{equation}
which summarizes the retrieved evidence. The retrieval history is then updated as $\mathcal{H}_t = \mathcal{H}_{t-1} \circ (q_t, o_t)$, and the LLM emits the next sub-query
\begin{equation}
    q_{t+1} \sim \pi_{\text{LLM}}(\cdot \mid \mathcal{H}_t)\text{.}
\end{equation}
The resulting next state is $s_{t+1} = (\mathcal{H}_t, q_{t+1})$.

Under this formulation, the state transition probability can be expressed as
\begin{equation}
    \resizebox{0.91\hsize}{!}{$
    \begin{aligned}
        P(s_{t+1} \mid s_t, a_t) &= \sum_{o_t} P(s_{t+1}, o_t \mid s_t, a_t) \\
        &= \sum_{o_t} P(o_t \mid s_t, a_t) \cdot P(s_{t+1} \mid s_t, a_t, o_t) \\
        &= \sum_{o_t} \underbrace{\pi_{\text{LLM}}(o_t \mid q_t, D_t)}_{\text{Observation Generation}} \cdot \underbrace{\pi_{\text{LLM}}(q_{t+1} \mid \mathcal{H}_t)}_{\text{Sub-query Generation}}\text{.}
    \end{aligned}
    $}
\end{equation}

\paragraph{Reward $r$.}
To optimize the retriever for final question answering performance, we use a sparse terminal reward:
\begin{equation}
    r_t = 
    \begin{cases}
        \mathrm{F1}(y, y^*), & t = T \\
        0, & t < T\text{,}
    \end{cases}
\end{equation}
where $y$ is the answer generated by conditioning the LLM on the full retrieval history $\mathcal{H}_T$, and $y^*$ is the ground-truth answer. The F1 score is computed as
\begin{equation}
    \mathrm{F1}(y, y^*)
    = \frac{2 \, |\mathcal{T}(y) \cap \mathcal{T}(y^*)|}
           {|\mathcal{T}(y)| + |\mathcal{T}(y^*)|}\text{,}
\end{equation}
where $\mathcal{T}(\cdot)$ denotes the set of tokens in a given text.

\subsection{Policy Parametrization} \label{sec:policy_formulation}
We parameterize the retriever as a stochastic policy $\pi_\theta$ that maps a state $s_t$ to a probability distribution over ordered lists of $k$ documents $D_t$, from which an action $a_t = D_t$ is sampled at each retrieval step.

To model this distribution, we view retrieval as sequential sampling without replacement from the corpus, which is modeled using the Plackett-Luce ranking model~\cite{plackett1975analysis}. Under this model, the probability of selecting an ordered document list $D_t = (d_t^1, \ldots, d_t^k)$ factorizes into a product of conditional selection probabilities
\begin{equation}
    \pi_\theta(a_t=D_t \mid s_t)
    = \prod_{i=1}^{k}
    \pi_\theta\!\left(d_t^i \mid s_t, d_t^{1:i-1}\right)\text{.}
\end{equation}

At each position $i$, the next document is sampled from the remaining candidate set $\mathcal{D} \setminus \{d_t^1, \ldots, d_t^{i-1}\}$ according to a softmax distribution
\begin{equation}
    \resizebox{0.91\hsize}{!}{$
    \begin{aligned}
        \pi_\theta\!\left(d_t^i \mid s_t, d_t^{1:i-1}\right)
        =
        \frac{\exp\!\left(f_\theta(s_t, d_t^i) / \alpha \right)}
        {\sum_{d \in \mathcal{D} \setminus \{d_t^j\}_{j=1}^{i-1}}
        \exp\!\left(f_\theta(s_t, d) / \alpha \right)}\text{,}
    \end{aligned}
    $}
    \label{eq:softmax_dist}
\end{equation}
where $\alpha > 0$ is a temperature parameter. The scoring function $f_\theta(s, d)$ represents the policy score of document $d$ under state $s$ and is defined as
\begin{equation}
    f_\theta(s, d)
    = \mathrm{sim}\!\left(\mathrm{enc}_\theta(s), \mathrm{enc}_\theta(d)\right)\text{,}
    \label{eq:score_func}
\end{equation}
where $\mathrm{enc}_\theta(\cdot)$ is a text encoder parameterized by $\theta$, and $\mathrm{sim}(\cdot,\cdot)$ denotes a similarity function (e.g., inner product).

\subsection{Policy Optimization} \label{sec:policy_opt}
Given the MDP formulation and the policy parametrization, our objective is to optimize the retriever policy to maximize the expected terminal reward,
\begin{equation}
    \max_{\theta} \; \mathbb{E}_{\tau \sim \pi_\theta} \big[ r_{|\tau|} \big]\text{,}
\end{equation}
where $\tau = (s_1, a_1, \ldots, s_{|\tau|}, a_{|\tau|})$ denotes a retrieval trajectory induced by the policy, and $r_{|\tau|}$ is the terminal reward defined in \S~\ref{sec:mdp_formulation}.

In our formulation, rewards are sparse and only observed at the end of a trajectory, while the state space is history-dependent and high-dimensional. Estimating accurate value functions in this setting is challenging and introduces additional model complexity. We therefore adopt Group Relative Policy Optimization (GRPO)~\cite{shao2024deepseekmath}, which optimizes the policy using relative outcome-based advantages without explicit value function estimation.

Specifically, for a given query $q_0 \sim P(Q)$, we sample a group of $G$ retrieval trajectories
$\{\tau_i\}_{i=1}^G$ from the old policy $\pi_{\theta_{\text{old}}}$. Each trajectory $\tau_i$ receives a terminal reward $r_i$, which is normalized within the group to obtain the advantage
\begin{equation}
    A_i
    =
    \frac{r_i - \mathrm{mean}(\mathbf{r})}{\mathrm{std}(\mathbf{r})},
    \quad
    \mathbf{r} = \{r_1,\ldots,r_G\}\text{.}
\end{equation}

The retriever policy is then optimized by maximizing the GRPO objective
\begin{equation}
    \resizebox{0.89\hsize}{!}{$
    \begin{aligned}
        \mathcal{J}_{\text{GRPO}}(\theta)
        =
        \mathbb{E}\Bigg[
        &\frac{1}{G}
        \sum_{i=1}^G
        \frac{1}{|\tau_i|}
        \sum_{t=1}^{|\tau_i|}
        \min\Bigg(
        \frac{\pi_\theta(a_{i,t} \mid s_{i,t})}
              {\pi_{\theta_{\text{old}}}(a_{i,t} \mid s_{i,t})} A_i, \\
        &
        \mathrm{clip}\!\left(
        \frac{\pi_\theta(a_{i,t} \mid s_{i,t})}
             {\pi_{\theta_{\text{old}}}(a_{i,t} \mid s_{i,t})},
        1-\epsilon,\,1+\epsilon
        \right) A_i
        \Bigg)
        \Bigg]\text{,}
    \end{aligned}
    $}
    \label{eq:grpo}
\end{equation}
where $\epsilon$ is the clipping parameter.

\subsection{Practical Training Approximations} \label{sec:approx}
The GRPO objective optimizes the retriever policy over the full document corpus. However, computing the policy probabilities in \eqref{eq:grpo} is prohibitively expensive for large-scale corpora, as it requires normalization over the entire corpus at each retrieval step and recomputation of document embeddings after each policy update. We therefore introduce the following approximations to enable efficient training.

\paragraph{Candidate Pool Restriction.}
At each retrieval step $t$, instead of normalizing over the full corpus $\mathcal{D}$, we first construct a candidate pool $\mathcal{C}_t \subset \mathcal{D}$ by retrieving the top-$K$ documents using the scoring function in \eqref{eq:score_func}, where $k < K \ll |\mathcal{D}|$. Policy sampling and probability normalization are then restricted to this candidate pool. Under this approximation, the conditional selection probability in \eqref{eq:softmax_dist} is given by
\begin{equation}
    \resizebox{0.89\hsize}{!}{$
    \begin{aligned}
        \pi_\theta\!\left(d_t^i \mid s_t, d_t^{1:i-1}\right)
        \approx
        \frac{\exp\!\left(f_\theta(s_t, d_t^i) / \alpha \right)}
        {\sum_{d \in \mathcal{C}_t \setminus \{d_t^j\}_{j=1}^{i-1}}
        \exp\!\left(f_\theta(s_t, d) / \alpha \right)}\text{.}
    \end{aligned}
    $}
\end{equation}

\paragraph{Document Encoder Freezing.}
To further reduce computational overhead, we freeze the document encoder during training and only update the state encoder. The scoring function in \eqref{eq:score_func} then takes the form
\begin{equation}
    f_\theta(s, d)
    =
    \mathrm{sim}\!\left(
    \mathrm{enc}_{\theta}(s),
    \mathrm{enc}(d)
    \right),
\end{equation}
where document embeddings $\mathrm{enc}(d)$ are precomputed and reused across policy updates, while the retriever adapts through learned state representations.

\section{Experiments}
The experiments aim at demonstrating:
(1) \app improves end-to-end question answering performance across different RAG pipelines, datasets, and retriever encoders (\S\ref{sec:compare_acc});
(2) the proposed RL-based fine-tuning and history-aware state representation both contribute positively to performance gains (\S\ref{sec:ablation_study});
and (3) \app achieves favorable training efficiency in terms of convergence speed, training time, and memory usage (\S\ref{sec:training_efficiency}).

\subsection{Experimental Setup}
We position \app as a \emph{plug-in} retriever optimization module compatible with existing RAG systems. To enable a fair assessment, we fix the RAG pipeline configurations—including LLMs, prompts, and interaction workflows—and evaluate performance gains solely by varying the retriever optimization strategy.

\paragraph{RAG Pipelines.}
We evaluate our method on two representative RAG pipelines to demonstrate its versatility.
(1) \emph{ReAct Agent}~\cite{yao2023react}: a ReAct-style pipeline using Qwen3-8B\footnote{Adopt checkpoint from \url{https://huggingface.co/Qwen/Qwen3-8B}.} as the LLM, where the retriever is invoked as an external tool.
(2) \emph{Search-R1}~\cite{jin2025search}: a pipeline with an RL-fine-tuned Qwen2.5-7B\footnote{Adopt checkpoint from \url{https://huggingface.co/PeterJinGo/SearchR1-nq_hotpotqa_train-qwen2.5-7b-em-ppo}.} LLM that interleaves reasoning and retrieval, with the retriever serving as its search engine.

\paragraph{Baselines.}
We compare against the following baselines under the same fixed RAG pipelines.
(1) \emph{Frozen Retriever}: the pre-trained retriever is used directly without adaptation, serving as a lower-bound reference.
(2) \emph{REPLUG}: a strong baseline based on the LSR method from REPLUG~\cite{shi2024replug}, which fine-tunes the retriever by matching its retrieval score distribution to the LLM's perplexity distribution over the top-$k$ documents.

\paragraph{Datasets \& Evaluation.}
To provide a comprehensive evaluation, we conduct experiments on both multi-hop and single-hop question answering benchmarks, using HotpotQA~\cite{yang2018hotpotqa} and Natural Questions (NQ)~\cite{kwiatkowski2019natural}, respectively. Following prior work~\cite{guan2025deeprag}, we report SQuAD~\cite{rajpurkar-etal-2016-squad} Exact Match (EM) and F1 scores as evaluation metrics.

\paragraph{Implementation.}
All experiments are conducted on a single node equipped with 8 NVIDIA A100 GPUs. Our method is implemented using the \texttt{TRL}~\cite{vonwerra2020trl} and \texttt{LlamaIndex}~\cite{Liu_LlamaIndex_2022} libraries, while REPLUG follows the \texttt{FedRAG}~\cite{Fajardo_fed-rag_2025} implementation, with all LLMs deployed via \texttt{vLLM}~\cite{kwon2023efficient}. Unless otherwise specified, we employ the 2018 Wikipedia dump~\cite{karpukhin2020dense} as the retrieval corpus, indexed in \texttt{Qdrant}~\cite{qdrant} using an INT8-quantized HNSW index ($m=16$, $\texttt{ef\_construct}=100$). We use Qwen3-Embedding-4B\footnote{Adopt checkpoint from \url{https://huggingface.co/Qwen/Qwen3-Embedding-4B}.} and -0.6B\footnote{Adopt checkpoint from \url{https://huggingface.co/Qwen/Qwen3-Embedding-0.6B}.} as retriever encoders, fine-tuned via LoRA~\cite{hu2022lora} ($r=32$, $\alpha=64$, dropout $=0.05$) using the AdamW~\cite{loshchilov2017decoupled} optimizer with FP16 precision, a constant learning rate of $1\times10^{-5}$ without warmup, a batch size of 16, and 100 training steps. For our method, we filter the training set to ensure non-zero reward variance under the initial policy across 8 rollouts and set the training retrieval top-$k$ to 3, with a temperature $\alpha=0.05$, a candidate pool size $K=30$, a GRPO group size $G=8$, and a clipping ratio $\epsilon=0.2$ (no KL regularization). At inference time, all methods use $k=3$ and are evaluated on the full, unfiltered development splits following the original dataset partitions for a fair comparison.

\subsection{Comparison of Accuracy} \label{sec:compare_acc}
\begin{table*}[t]
    \centering
    \small
    \setlength{\tabcolsep}{4.5pt}
    \caption{End-to-end QA performance (EM/F1; higher is better) across two RAG agents (ReAct vs. Original), two datasets (HotpotQA, NQ), and two retriever encoders (Qwen3-Embedding-4B, Qwen3-Embedding-0.6B). For each base agent, we compare the frozen-retriever baseline (ReAct Agent / Search-R1), a plug-in retrieval variant (REPLUG), and our method \app applied within the same agent pipeline. \textbf{Bold} denotes the best result in each column, and \underline{underline} denotes the second-best. \textit{Gains} reports the relative improvement (\%) of \app over the strongest non-\app baseline under the same base agent for that column.}

    \label{tab:compare_acc}
    \begin{tabular}{clcccccccccc}
        \toprule
        \multirow{3.7}{*}{\textbf{Base Agent}} & \multirow{3.7}{*}{\textbf{Approach}} & \multicolumn{4}{c}{\textbf{Qwen3-Embedding-4B}} & \multicolumn{4}{c}{\textbf{Qwen3-Embedding-0.6B}} & \multicolumn{2}{c}{\multirow{2.35}{*}{\textbf{Average}}} \\
        \cmidrule(lr){3-6} \cmidrule(lr){7-10}
          & & \multicolumn{2}{c}{\textbf{HotpotQA}} & \multicolumn{2}{c}{\textbf{NQ}} & \multicolumn{2}{c}{\textbf{HotpotQA}} & \multicolumn{2}{c}{\textbf{NQ}} & \multicolumn{2}{c}{} \\
        \cmidrule(lr){3-4} \cmidrule(lr){5-6} \cmidrule(lr){7-8} \cmidrule(lr){9-10} \cmidrule(lr){11-12}
        & & EM & F1 & EM & F1 & EM & F1 & EM & F1 & EM & F1 \\
        \midrule
        \multirow{4}{*}{ReAct} & Original & 31.28 & 40.42 & \underline{32.82} & \underline{43.86} & \underline{28.74} & \underline{37.47} & 30.10 & 40.46 & 30.73 & 40.55 \\
        & REPLUG & \underline{31.91} & \underline{40.91} & 32.60 & 43.77 & 28.53 & 37.04 & \underline{30.12} & \underline{40.58} & \underline{30.79} & \underline{40.57} \\
        \cmidrule{2-12}
        & \textbf{\app} (Ours) & \textbf{32.34} & \textbf{41.55} & \textbf{33.46} & \textbf{44.56} & \textbf{29.60} & \textbf{38.14} & \textbf{30.16} & \textbf{40.72} & \textbf{31.39} & \textbf{41.24} \\
        & \textit{Gains} & +1.35\% & +1.56\% & +1.95\% & +1.60\% & +3.01\% & +1.76\% & +0.11\% & +0.36\% & +1.61\% & +1.32\% \\
        \midrule
        \multirow{4}{*}{Search-R1} &  Original & 40.04 & 50.48 & 43.91 & \underline{53.61} & \underline{37.88} & \underline{47.81} & \underline{40.92} & \underline{50.08} & 40.69 & 50.50 \\
        & REPLUG & \textbf{40.39} & \underline{50.90} & \underline{43.95} & 53.52 & 37.66 & 47.67 & \textbf{41.12} & \textbf{50.27} & \underline{40.78} & \underline{50.59} \\
        \cmidrule{2-12}
        & \textbf{\app} (Ours) & \underline{40.30} & \textbf{51.02} & \textbf{44.92} & \textbf{54.45} & \textbf{38.34} & \textbf{48.63} & 40.87 & 50.07 & \textbf{41.11} & \textbf{51.04} \\
        & \textit{Gains} & -0.23\% & +0.24\% & +2.21\% & +1.57\% & +1.21\% & +1.72\% & -0.61\% & -0.40\% & +0.64\% & +0.78\% \\
        \bottomrule
    \end{tabular}%
\end{table*}
\begin{figure*}[t]
    \centering
    \begin{subfigure}[c]{0.245\textwidth}
        \centering
        \includegraphics[width=\linewidth]{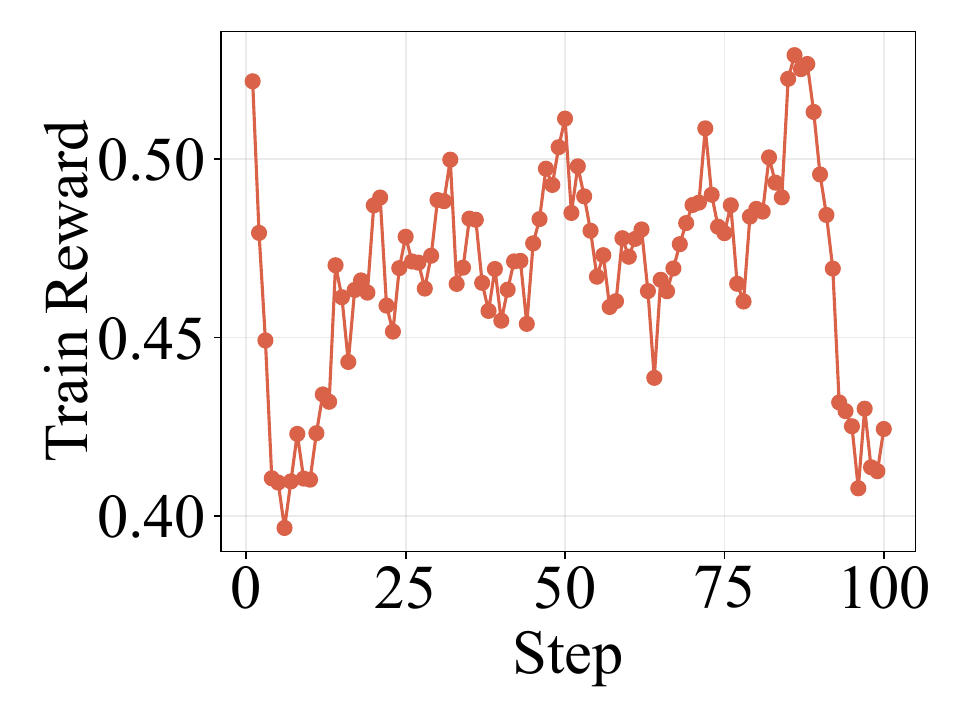}
        \caption{React Agent 4B}
        \label{fig:train_reward_react_agent_4b}
    \end{subfigure}
    \begin{subfigure}[c]{0.245\textwidth}
        \centering
        \includegraphics[width=\linewidth]{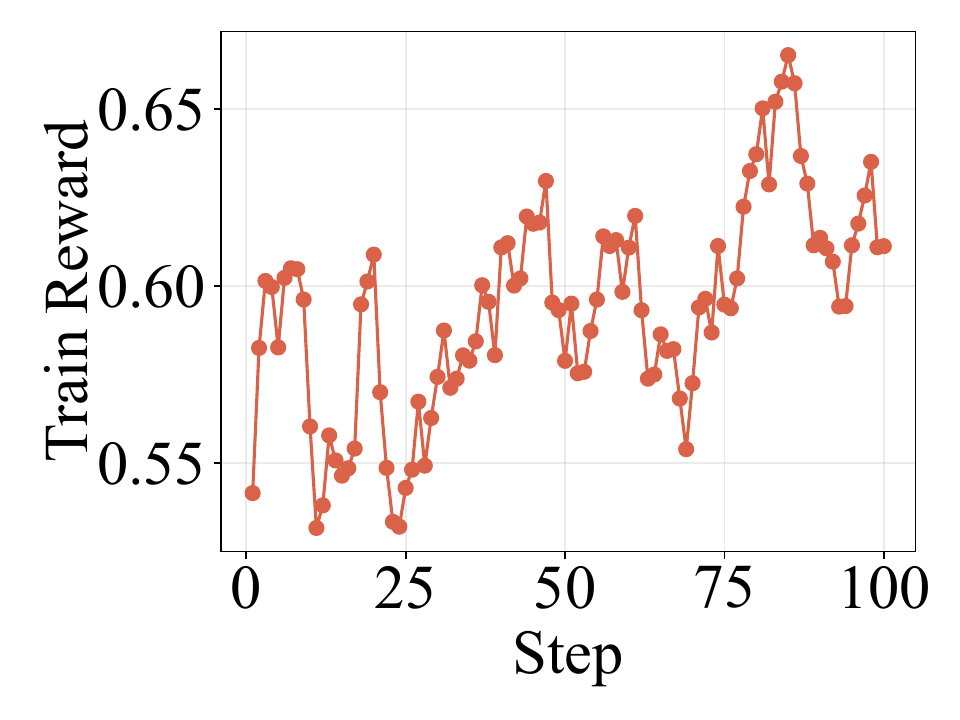}
        \caption{Search-R1 4B}
        \label{fig:train_reward_search_r1_4b}
    \end{subfigure}
    \begin{subfigure}[c]{0.245\textwidth}
        \centering
        \includegraphics[width=\linewidth]{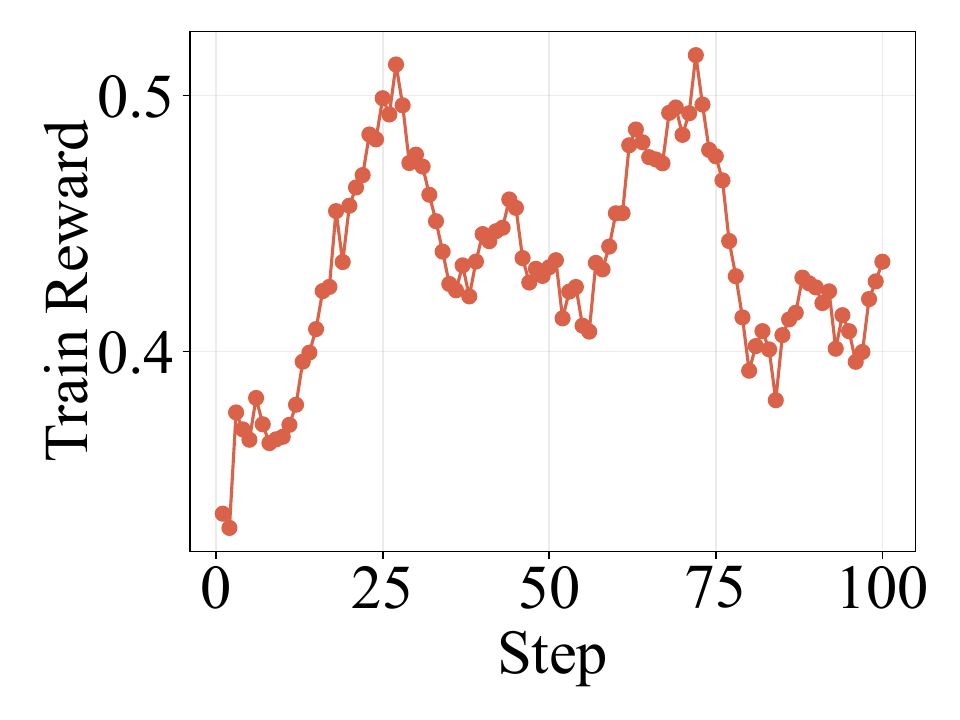}
        \caption{React Agent 0.6B}
        \label{fig:train_reward_react_agent_0.6b}
    \end{subfigure}
    \begin{subfigure}[c]{0.245\textwidth}
        \centering
        \includegraphics[width=\linewidth]{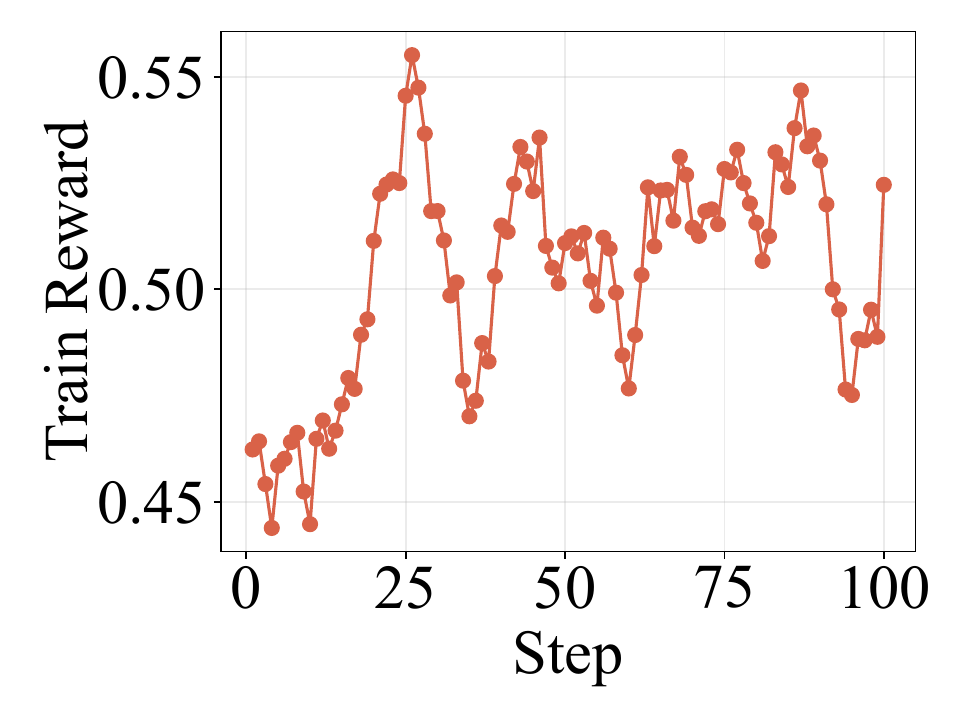}
        \caption{Search-R1 0.6B}
        \label{fig:train_reward_search_r1_0.6b}
    \end{subfigure}
    \caption{Training reward trajectories on the HotpotQA dataset. Subplot titles denote the specific RAG pipeline and retriever encoder used. All curves are smoothed using Exponential Moving Average (EMA) with a window size of 8.}
    \label{fig:train_reward}
\end{figure*}

Table~\ref{tab:compare_acc} reports the end-to-end question answering performance across different RAG pipelines, datasets, and retriever encoders, while Figure~\ref{fig:train_reward} depicts the corresponding training reward trajectories on HotpotQA. We analyze the results from four complementary perspectives.

\paragraph{Comparison with Frozen Retrievers.}
From Table~\ref{tab:compare_acc}, our approach improves performance in 18 out of 20 reported metrics compared to the Frozen Retriever baseline, demonstrating consistent gains across diverse experimental settings. We attribute these improvements to end-to-end reinforcement learning, which aligns the retriever with the LLM and enables more effective coordination between retrieval and reasoning components.

\paragraph{Superiority over Supervised Fine-Tuning.}
From Table~\ref{tab:compare_acc}, our method also outperforms REPLUG in 17 out of 20 metrics. While REPLUG optimizes proxy objectives that approximate LLM preferences, such objectives do not necessarily translate into improved end-to-end question answering performance. In contrast, our approach directly optimizes the downstream task reward, reducing the mismatch between training objectives and inference-time behavior and allowing the retriever to learn document selection strategies that better support the overall reasoning pipeline.

\paragraph{Training Dynamics.}
Figure~\ref{fig:train_reward} illustrates the training reward trajectories of \app. We observe an initial decrease in reward followed by a steady increase. This early decline is likely caused by a distribution shift from the history-aware state representation, which differs from the retriever's pretraining regime. After an adaptation phase, rewards increase consistently, indicating that the retriever learns to exploit historical context to improve retrieval decisions.

\paragraph{Analysis of Performance Variations.}
Despite the overall gains, several trends are worth noting. First, reward trajectories exhibit oscillations during training (Figure~\ref{fig:train_reward}), which can be attributed to sparse terminal rewards and the off-policy approximations introduced for scalability (see \S\ref{sec:approx}). Second, improvements are more pronounced on the multi-hop HotpotQA dataset than on the single-hop NQ benchmark, and when using the larger 4B retriever encoder compared to the 0.6B model. These observations suggest that history-aware state representations are most beneficial when retrieval decisions depend on longer reasoning contexts, and that sufficient model capacity is required to effectively leverage such contextual information.

\subsection{Ablation Study} \label{sec:ablation_study}
\begin{table*}[t]
    \centering
    \caption{Ablation study on the ReAct Agent pipeline across two datasets and two retriever encoders. ``w/o History'' and ``w/o RL'' refer to removing the history-aware state space and the RL-based retriever fine-tuning from \app, respectively. Notations follow Table~\ref{tab:compare_acc}.}
    \label{tab:ablation_study}
    \begin{tabular}{lcccccccccc}
        \toprule
        \multicolumn{1}{c}{\multirow{3.7}{*}{\makecell{\textbf{Approach based on} \\ \textbf{ReAct Agent}}}} & \multicolumn{4}{c}{\textbf{Qwen3-Embedding-4B}} & \multicolumn{4}{c}{\textbf{Qwen3-Embedding-0.6B}} & \multicolumn{2}{c}{\multirow{2.35}{*}{\textbf{Average}}} \\
        \cmidrule(lr){2-5} \cmidrule(lr){6-9}
         & \multicolumn{2}{c}{\textbf{HotpotQA}} & \multicolumn{2}{c}{\textbf{NQ}} & \multicolumn{2}{c}{\textbf{HotpotQA}} & \multicolumn{2}{c}{\textbf{NQ}} & \multicolumn{2}{c}{} \\
        \cmidrule(lr){2-3} \cmidrule(lr){4-5} \cmidrule(lr){6-7} \cmidrule(lr){8-9} \cmidrule(lr){10-11}
         & EM & F1 & EM & F1 & EM & F1 & EM & F1 & EM & F1 \\
        \midrule
        Original & 31.28 & 40.42 & 32.82 & 43.86 & 28.74 & 37.47 & 30.10 & 40.46 & 30.73 & 40.55 \\
        \app w/o History    & \underline{31.56} & \underline{40.74} & 32.68 & 43.86 & \underline{29.37} & \underline{37.99} & \textbf{30.19} & \underline{40.48} & \underline{30.95} & \underline{40.77} \\
        \app  w/o RL       & 31.42 & 40.63 & \underline{32.98} & \underline{44.14} & 27.52 & 35.74 & 29.80 & 40.20 & 30.43 & 40.18 \\ 
        \midrule
        \app       & \textbf{32.34} & \textbf{41.55} & \textbf{33.46} & \textbf{44.56} & \textbf{29.60} & \textbf{38.14} & \underline{30.16} & \textbf{40.72} & \textbf{31.39} & \textbf{41.24} \\
        \bottomrule
    \end{tabular}%
\end{table*}
\begin{figure}[t]
    \centering
    \begin{subfigure}[t]{0.49\linewidth}
        \centering
        \includegraphics[width=\linewidth]{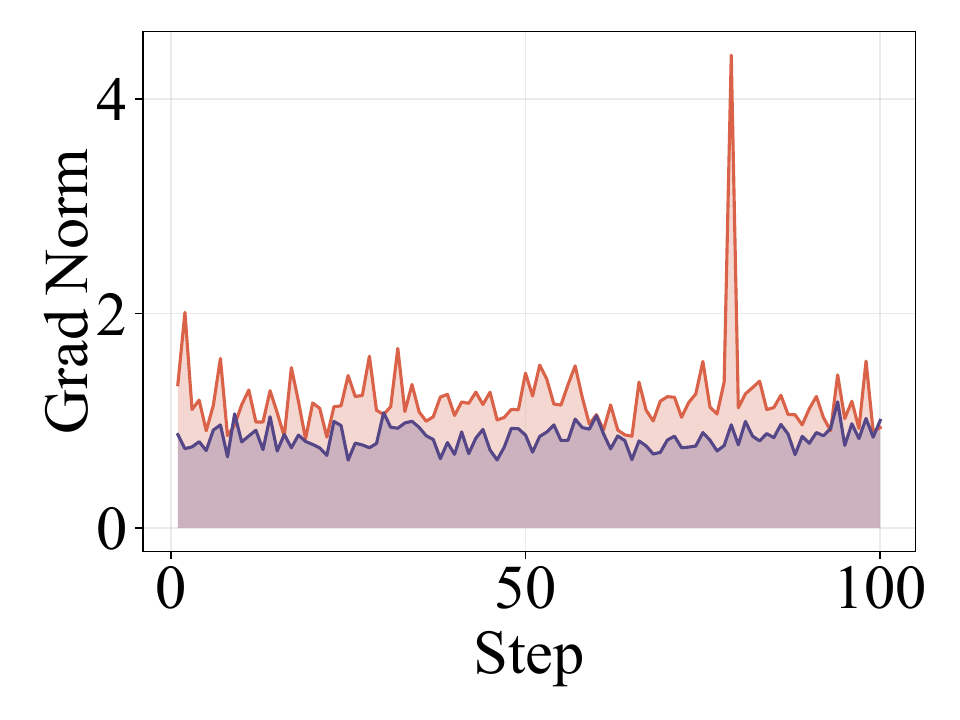}
        \caption{4B}
        \label{fig:grad_norm_4b}
    \end{subfigure}
    \begin{subfigure}[t]{0.49\linewidth}
        \centering
        \includegraphics[width=\linewidth]{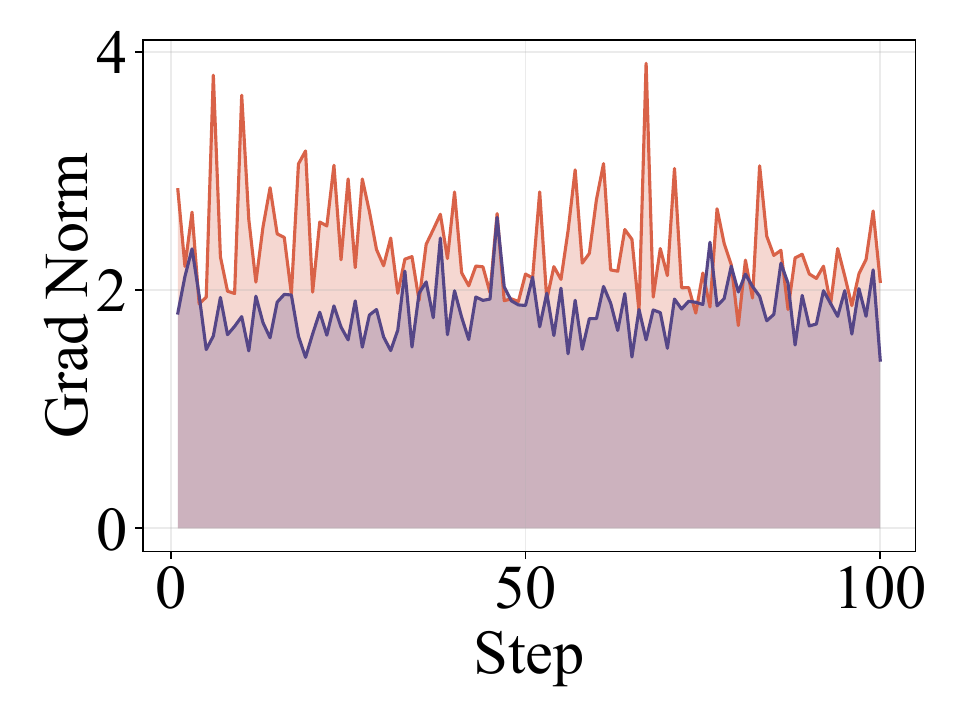}
        \caption{0.6B}
        \label{fig:grad_norm_0.6b}
    \end{subfigure}

    \vspace{0.1cm}
    \begin{subfigure}[t]{0.5\linewidth}
        \centering
        \includegraphics[width=\linewidth]{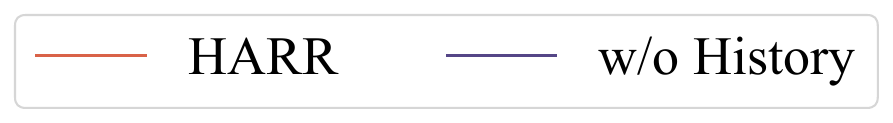}
    \end{subfigure}
    \caption{Gradient norm trajectories on HotpotQA using the ReAct Agent pipeline. Subplot titles denote the retriever encoder size.}
    \label{fig:grad_norm}
\end{figure}

We conduct ablation studies on the ReAct Agent pipeline to analyze the contributions of the proposed history-aware state representation and the RL-based retriever fine-tuning within \app. Table~\ref{tab:ablation_study} reports the results across two datasets and two retriever encoder sizes, while Figure~\ref{fig:grad_norm} illustrates the corresponding gradient norm trajectories.

\paragraph{Effect of RL and History-Aware State.}
Table~\ref{tab:ablation_study} demonstrates that both proposed components are essential for achieving optimal performance. Specifically, removing the history-aware state degrades performance in 9 out of 10 metrics, while excluding the RL-based optimization leads to drops across all 10 metrics. The notably larger degradation caused by removing RL identifies it as the primary driver of performance gains, while the history-aware state serves as a critical complementary module.

\paragraph{RL versus History-Aware State in Isolation.}
We further analyze the behavior of each component in isolation. When employing the history-aware state without RL-based fine-tuning, the average performance fails to surpass that of the frozen retriever. We attribute this to the distribution shift introduced by the augmented state, which the pre-trained encoder cannot effectively leverage without end-to-end adaptation. Conversely, applying RL without the history-aware state still yields improvements over the frozen baseline, indicating that our RL framework is robust and capable of enhancing performance even in the presence of state aliasing.

\paragraph{Impact of Model Capacity.}
The relative contribution of each component varies with retriever capacity. For the smaller 0.6B encoder, adding RL-based fine-tuning on top of the frozen retriever yields larger marginal gains. This suggests that smaller models may initially suffer from a larger gap in matching LLM preferences, thereby benefiting more from the RL-driven optimization. In contrast, for the larger 4B encoder, introducing the history-aware state on top of RL leads to more pronounced improvements, indicating that higher-capacity models are better equipped to exploit historical context for informed retrieval decisions.

\paragraph{Gradient Analysis.}
Finally, Figure~\ref{fig:grad_norm} provides insight into the optimization dynamics. Compared to the variant trained without history, \app exhibits consistently larger gradient norms throughout training. This observation is consistent with our hypothesis that the history-aware state mitigates state aliasing, thereby reducing reward ambiguity and yielding stronger learning signals for policy optimization.

\subsection{Training Efficiency} \label{sec:training_efficiency}
\begin{table}[h]
    \centering
    \setlength{\tabcolsep}{4pt}
    \caption{Training efficiency analysis. We report the peak GPU memory usage (Mem) and total training time (Time) for a single experiment run across varying RAG pipelines, datasets, and retriever encoders. All models are fine-tuned with DP=1, TP=1, gradient checkpointing enabled, and no gradient accumulation.}
    \label{tab:training_efficiency}
    \begin{tabular}{lcccc}
        \toprule
        \multirow{3.7}{*}{RAG Pipeline} & \multicolumn{4}{c}{\textbf{Qwen3-Embedding-4B}} \\
        \cmidrule(lr){2-5}
         & \multicolumn{2}{c}{\textbf{HotpotQA}} & \multicolumn{2}{c}{\textbf{NQ}} \\
        \cmidrule(lr){2-3} \cmidrule(lr){4-5}
         & Mem/GB & Time/h & Mem/GB & Time/h \\
        \midrule
        ReAct Agent & 61.70 & 3.10 & 31.42 & 3.14 \\
        Search-R1   & 65.82 & 1.36 & 41.72 & 1.12 \\
        \toprule
        \multirow{3.7}{*}{RAG Pipeline} & \multicolumn{4}{c}{\textbf{Qwen3-Embedding-0.6B}} \\
        \cmidrule(lr){2-5}
         & \multicolumn{2}{c}{\textbf{HotpotQA}} & \multicolumn{2}{c}{\textbf{NQ}} \\
        \cmidrule(lr){2-3} \cmidrule(lr){4-5}
         & Mem/GB & Time/h & Mem/GB & Time/h \\
        \midrule
        ReAct Agent & 26.84 & 2.40 & 17.70 & 2.00 \\
        Search-R1   & 31.48 & 2.42 & 16.47 & 1.51 \\
        \bottomrule
    \end{tabular}%
\end{table}
Since \app exclusively fine-tunes the retriever encoder while keeping the LLM and inference workflow unchanged, it introduces no additional inference overhead; therefore, we focus on training efficiency in terms of convergence speed, wall-clock training time, and memory usage.
As shown by the training reward curves in Figure~\ref{fig:train_reward}, RL-based retriever optimization converges rapidly, typically within 100 training steps across different settings.
Consistent with this observation, Table~\ref{tab:training_efficiency} shows that the total training time of a single experiment is at most around three hours, which is substantially lower than the cost of fine-tuning LLMs, often reported to require over ten hours even with comparable hardware~\cite{sun2025improvingdataefficiencyllm}.
In terms of memory consumption, training the 0.6B retriever requires less than 32~GB GPU memory, making it feasible on consumer-grade GPUs.
While the 4B retriever incurs higher memory usage under our default configuration, this is mainly due to the absence of aggressive memory optimization; in practice, techniques such as gradient accumulation can be applied to reduce peak memory footprint without significantly increasing training time.
Overall, these results demonstrate that \app offers a favorable efficiency--effectiveness trade-off for retriever optimization in RAG systems.
\section{Conclusion}
\label{sec:conclusion}
This work targets the core mismatch between supervised retriever objectives and end-to-end QA in RAG, and proposes \app: a history-aware RL framework that fine-tunes dense retrievers by (i) replacing deterministic top-$k$ with stochastic sampling to expose action distributions and (ii) encoding retrieval history in the state to reduce aliasing in multi-hop reasoning. Across datasets, RAG pipelines, and retriever scales, \app consistently improves end-to-end RAG performance, positioning RL as a retriever-centric alternative that optimizes retrieval behavior directly for downstream task reward rather than proxy relevance signals.

\section*{Impact Statement}
This paper presents work whose goal is to advance the field of Machine
Learning. There are many potential societal consequences of our work, none
which we feel must be specifically highlighted here.




\nocite{langley00}

\bibliography{reference}
\bibliographystyle{icml2026}

\newpage
\appendix
\onecolumn

\end{document}